\def\paperID{0000}
\def\confName{CVPR}
\def\confYear{2026}

\def\paperTitle{SpeechParaling-Bench: A Comprehensive Benchmark for\\
Paralinguistic-Aware Speech Generation}

\def\authorBlock{
   Ruohan Liu$^{1\ast}$, ~Shukang Yin$^{1}$\thanks{Equal contribution.}\hspace{1.1mm}, ~Tao Wang$^{1}$, ~Dong Zhang$^{2}$, 
   ~Weiji Zhuang$^{2}$, ~Shuhuai Ren$^{2}$,
   \\Ran He$^{1}$, ~Caifeng Shan$^{1}$, ~Chaoyou Fu$^{1}$\thanks{Corresponding author.} \\[0.15em]
$^1$Nanjing University, $^2$Xiaomi \\
    {\tt\small ruohanliu998@gmail.com, bradyfu24@gmail.com}\\[0.2em]
Project page: \href{https://speechparaling-bench.github.io/}{speechparaling-bench.github.io}
}

\newif\ifreview 
\newif\ifarxiv \newcommand{\arxiv}{\arxivtrue}
\newif\ifcamera 
\newif\ifrebuttal 

\arxiv 

\pdfoutput=1
\documentclass[10pt,twocolumn,letterpaper]{article}
\ifreview \usepackage[review]{cvpr} \fi
\ifarxiv \usepackage[pagenumbers]{cvpr} \fi
\ifrebuttal \usepackage[rebuttal]{cvpr} \fi
\ifcamera \usepackage{cvpr} \fi

\usepackage[T1]{fontenc}  
\usepackage{graphicx}	
\usepackage{amsmath}	
\usepackage{amssymb}	
\usepackage{booktabs}
\usepackage{times}
\usepackage{microtype}
\usepackage{epsfig}
\usepackage{caption}
\usepackage{float}
\usepackage{placeins}
\usepackage{color, colortbl}
\usepackage{stfloats}
\usepackage{enumitem}
\usepackage{tabularx}
\usepackage{xstring}
\usepackage{multirow}
\usepackage{xspace}
\usepackage{url}
\usepackage{subcaption}
\usepackage{xcolor}
\usepackage[hang,flushmargin]{footmisc}

\usepackage{pifont}
\usepackage{bbding}
\usepackage{fontawesome5} 
\usepackage{makecell}    

\usepackage[most]{tcolorbox}
\usepackage{adjustbox}
\usepackage{textcomp}
\usepackage{changepage}

\ifcamera \usepackage[accsupp]{axessibility} \fi

\ifarxiv  \fi

\newcommand{\R}[1]{{%
    \textbf{%
        \ifstrequal{#1}{1}{\textcolor{red}{R#1}}{%
        \ifstrequal{#1}{2}{\textcolor{blue}{R#1}}{%
        \ifstrequal{#1}{3}{\textcolor{magenta}{R#1}}{%
        \ifstrequal{#1}{4}{\textcolor{teal}{R#1}}{%
                           \textcolor{cyan}{R#1}%
        }}}}%
    }%
}}

\newcommand{\name}[0]{\textsc{SpeechParaling-Bench}\xspace}
\newcommand{\rawname}[0]{SpeechParaling-Bench\xspace}

\newcommand{\tablestyle}[2]{\setlength{\tabcolsep}{#1}\renewcommand{\arraystretch}{#2}}
\newcommand{\shline}{\specialrule{.1em}{.05em}{.05em}}

\newcommand{\cmark}{\textcolor{Green}{\ding{51}}}
\newcommand{\xmark}{\textcolor{Red}{\ding{55}}}

\definecolor{gold}{HTML}{FFD700}
\definecolor{silver}{HTML}{C0C0C0} 
\definecolor{bronze}{HTML}{A0522D}

\newcommand{\best}[1]{\textbf{#1}\rlap{\,\textcolor{gold}{\raisebox{-0.1ex}{\faMedal}}}}

\newcommand{\second}[1]{\underline{#1}\rlap{\,\textcolor{silver}{\raisebox{-0.1ex}{\faMedal}}}}

\newcommand{\third}[1]{#1\rlap{\,\textcolor{bronze}{\raisebox{-0.1ex}{\faMedal}}}}

\usepackage[outline]{contour}
\contourlength{0.8pt}

\definecolor{ourline}{rgb}{1.0, 0.8, 0.6}
\definecolor{bulbcolor}{rgb}{1.0, 0.9, 0.0}
\definecolor{lightgray}{gray}{0.9}

\definecolor{expressive_color}{HTML}{add5d9}
\definecolor{prosodic_color}{HTML}{fab970}
\definecolor{acoustic_color}{HTML}{cbc8c0}

\usepackage{CJKutf8}

\newtcolorbox[auto counter]{takeaway}[1][]{
    colback=ourline!15,      
    colframe=ourline!95!red, 
    coltitle=black,          
    fonttitle=\bfseries,     
    title={\contour{black}{\textcolor{bulbcolor}{\faLightbulb}}\ \ Key Takeaway~\thetcbcounter}, 
    #1
}

\lstdefinelanguage{json}{
    basicstyle=\ttfamily\small, 
    numbers=none, 
    showstringspaces=false, 
    breaklines=true, 
    morestring=[b]",
    morecomment=[l]{//},
    morecomment=[s]{/*}{*/},
    morekeywords={true,false,null}
}  

\usepackage{xr-hyper}

\makeatletter
\newcommand*{\addFileDependency}[1]{
  \typeout{(#1)}
  \@addtofilelist{#1}
  \IfFileExists{#1}{}{\typeout{No file #1.}}
}

\makeatother
\newcommand*{\myexternaldocument}[1]{
    \externaldocument{#1}
    \addFileDependency{#1.tex}
    \addFileDependency{#1.aux}
}

\definecolor{cvprblue}{rgb}{0.21,0.49,0.74}
\usepackage[pagebackref,breaklinks,colorlinks,allcolors=cvprblue]{hyperref}
\usepackage[capitalize]{cleveref}
\crefname{section}{Sec.}{Secs.}
\crefname{table}{Table}{Tables}
\crefname{figure}{Fig.}{Figs.}

\ifarxiv \crefname{appendix}{App.}{Apps.}
\else \crefname{appendix}{Suppl.}{Suppls.} \fi

\unless\ifarxiv \myexternaldocument{_supplementary} \fi

\begin{document}
\title{\paperTitle}
\author{\authorBlock}
\maketitle

\begin{abstract}
Paralinguistic cues are essential for natural human-computer interaction, yet their evaluation in Large Audio-Language Models (LALMs) remains limited by coarse feature coverage and the inherent subjectivity of assessment. 
To address these challenges, we introduce \textbf{SpeechParaling-Bench}, a comprehensive benchmark for paralinguistic-aware speech generation. It expands existing coverage from fewer than 50 to \textbf{over 100 fine-grained features}, supported by more than 1,000 English-Chinese parallel speech queries, and is organized into three progressively challenging tasks: fine-grained control, intra-utterance variation, and context-aware adaptation. 
To enable reliable evaluation, we further develop \textbf{a pairwise comparison pipeline}, in which candidate responses are evaluated against a fixed baseline by an LALM-based judge. By framing evaluation as relative preference rather than absolute scoring, this approach mitigates subjectivity and yields more stable and scalable assessments without costly human annotation. 
Extensive experiments reveal substantial limitations in current LALMs. Even leading proprietary models struggle with comprehensive static control and dynamic modulation of paralinguistic features, while failure to correctly interpret paralinguistic cues accounts for 43.3\% of errors in situational dialogue. 
These findings underscore the need for more robust paralinguistic modeling toward human-aligned voice assistants.
\end{abstract}

\section{Introduction}
\label{sec:intro}

\begin{figure}[t!]
    \centering
    \setlength{\abovecaptionskip}{3mm}
    \setlength{\belowcaptionskip}{-4mm}
    \includegraphics[width=0.98\linewidth]{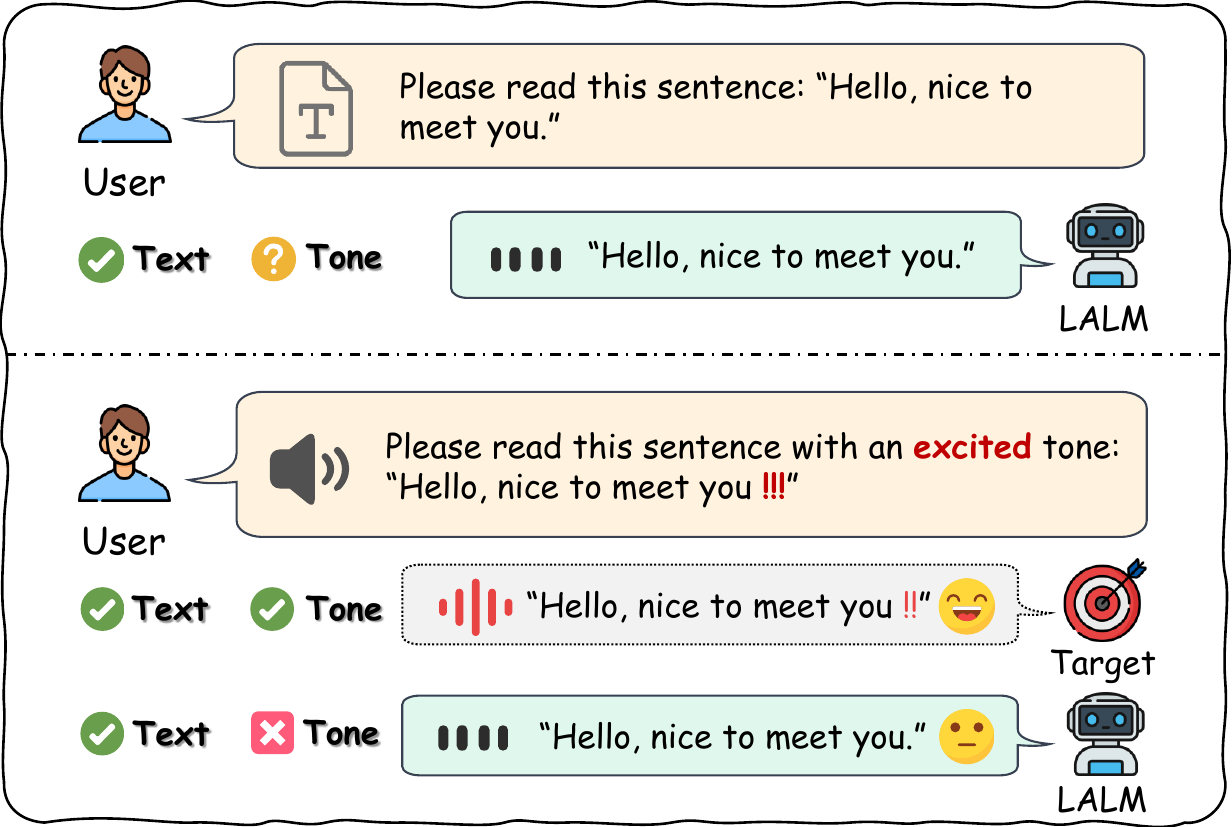} 
    \caption{\textbf{Comparison between traditional speech generation and paralinguistic-aware speech generation.}
    While traditional benchmarks (Top) focus on text-to-speech consistency, the latter (Bottom) requires the model to synthesize not only linguistic content but also non-verbal features (\eg, tone).}
    \label{fig:teaser}
\end{figure}

Recent years have witnessed the rise of Large Audio Language Models (LALMs)~\cite{chatgpt-voice, qwen2p5-omni, MLLM-survey}.
Different from the traditional audio modeling approach that tackles each audio processing task, \eg, speech recognition~\cite{chan2016listen,deep-speech,deep-speech-2}, emotion recognition~\cite{wav2vec,wav2vec-2}, and speech synthesis~\cite{tacotron,fastspeech-2}, LLM-driven audio modeling enables the emergence of audio foundation models~\cite{kimi-audio,mimo-audio}, which provide versatile task support with a unified I/O interface. 
Moreover, empowered by the strong language proficiency of the LLM backbone, new applications such as real-time spoken dialogue~\cite{glm-4-voice,moshi,chatgpt-voice} have become a reality.
Notably, frontier models like ChatGPT-Audio~\cite{chatgpt-voice} and Doubao Voice~\cite{doubao-voice} have demonstrated preliminary capabilities in paralinguistic-aware speech generation, mimicking human speaking styles and tones and facilitating more natural user interactions. 
Yet despite extensive evaluations of general audio tasks, assessments of competence in this important capability remain limited.
As shown in~\cref{fig:teaser}, proficiency in paralinguistic-aware speech generation requires not only the correct generation of linguistic content but also the accurate expression of non-verbal aspects such as speaking styles and tones.

\begin{figure*}[t!]
    \centering
    \setlength{\abovecaptionskip}{3mm}
    \setlength{\belowcaptionskip}{-4mm}
    \includegraphics[width=0.9\linewidth]{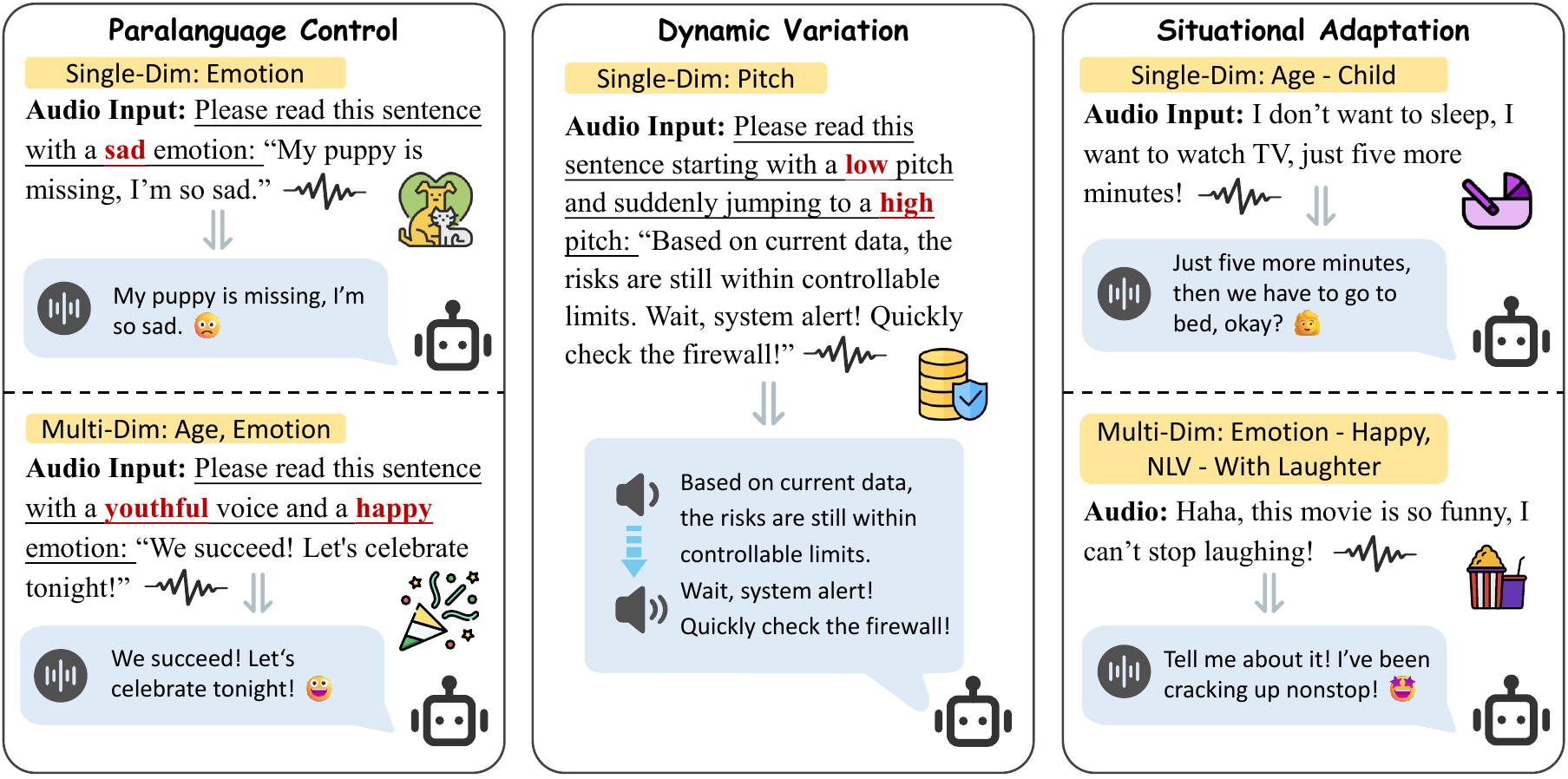}
    \caption{\textbf{Data samples from \rawname.} 
    Our evaluation covers three tasks critical for paralinguistic-aware speech generation: 
    (1) \textbf{Paralanguage Control}: tests the LALM's ability to generate audio with specific paralinguistic features; 
    (2) \textbf{Dynamic Variation}: assesses the capability to modulate paralinguistic features; and 
    (3) \textbf{Situational Adaptation}: evaluates the paralinguistic alignment between LALMs and users, where, unlike the former two, there is no standard answer for content.
    Each sample consists of an audio query paired with paralinguistic annotations.
    Single/Multi-Dim: Single-/Multi-dimension. NLV: Non-Linguistic Vocalizations.}
    \label{fig:sample}
\end{figure*}

To fill this gap, we introduce SpeechParaling-Bench, a comprehensive evaluation suite for paralinguistic-aware speech generation, featuring: 
\textbf{(1)} Broader paralinguistic feature coverage. 
Compared to existing benchmarks that typically cover fewer than 50 features, our benchmark expands the scope to over 100 distinct paralinguistic features, comprising more than 1,000 English–Chinese parallel speech queries curated via our custom data pipeline.
\textbf{(2)} Specialized task design. 
As shown in~\cref{fig:sample}, we structure the evaluation around progressive skill types, \textit{Paralanguage Control}, \textit{Dynamic Variation}, and \textit{Situational Adaptation}, ranging from controlled generation to context-aware adaptability, with a focus on real-world utility.
\textbf{(3)} Enhanced evaluation pipeline. 
To address the inherent subjectivity of paralinguistic evaluation, we adopt a pairwise comparison framework that evaluates candidate responses against a fixed baseline. 
By reducing the task to relative preference rather than absolute scoring, this approach yields more stable and reliable assessments while remaining efficient and scalable.

Through extensive evaluations on leading LALMs, we find that:
(1) Achieving a comprehensive and accurate control across various paralinguistic dimensions is still challenging;
(2) Dynamic regulation of paralinguistic features is a common bottleneck;
(3) Failing to understand the paralinguistic cues embedded in user speech is a major reason (accounting for 43.3\%) for failure in situational dialogue.

Overall, our contributions are threefold:

\begin{itemize}
    \item \textbf{A comprehensive benchmark:} We introduce a benchmark that expands paralinguistic coverage from fewer than 50 to over 100 fine-grained features, supported by more than 1,000 English–Chinese parallel speech queries. The benchmark is structured into three complementary tasks—fine-grained control, intra-utterance variation, and context-aware adaptation—to capture paralinguistic abilities from static to contextual settings.
    
    \item \textbf{A pairwise evaluation pipeline:} We propose an automated pairwise evaluation framework that compares candidate responses against a fixed baseline. By reformulating evaluation as relative preference rather than absolute scoring, the approach mitigates the inherent subjectivity of paralinguistic assessment, resulting in more stable and scalable evaluation without costly human annotation.
    
    \item \textbf{Empirical insights:} Through extensive experiments, we identify key limitations of current LALMs, including weak dynamic modulation and difficulty in capturing contextual paralinguistic cues, with misinterpretation of such cues accounting for a substantial portion of errors. These findings highlight critical bottlenecks for building more natural and human-aligned voice assistants.
\end{itemize}

\section{Related Work}
\label{sec:related}

\subsection{Large Audio-Language Models} 
Recent years have witnessed the rapid emergence of Large Audio-Language Models (LALMs) and their applications in real-world scenarios~\cite{step-audio-2, kimi-audio, mimo-audio}.
Equipped with exceptional reasoning and fluent speech-generation capabilities, these models facilitate seamless, colloquial dialogue experiences.
In the commercial domain, frontier models like GPT Audio~\cite{gpt0828} and Gemini Audio~\cite{gemini-audio} exemplify end-to-end multimodal understanding, offering native support for audio streaming and expressive responses rich in emotion and prosody.
Similarly, the Doubao Realtime Voice Model~\cite{doubao-voice} is tailored for the Chinese linguistic context, exhibiting extraordinary naturalness and ultra-low interaction latency. 
Conversely, the open-source community focuses on democratizing these capabilities through efficient architectures.
For instance, LLaMA-Omni~\cite{llama-omni, llama-omni2} and Freeze-Omni~\cite{freeze-omni} propose efficient fine-tuning techniques to integrate audio modalities without compromising the textual proficiency of the original LLM backbones.
Qwen-Omni series~\cite{qwen2p5-omni, qwen3-omni} adopt a Thinker-Talker architecture, achieving a balance between reasoning capabilities and generation efficiency under a relatively small parameter scale.

\begin{table}[t!]
    \centering
    \small
    \tablestyle{18pt}{1.1}
    \vspace{-2mm}
    \caption{\textbf{Key statistics of \rawname.}}
    
    \begin{tabular}{l r}
    \shline
    \textbf{Statistic} & \textbf{Number} \\
    \hline
    Samples & 1,001 \\
    Dimensions & 13 \\
    Features & 104 \\
    \hline
    Avg. Text Length & \\
    \hspace{3mm} - \textit{English Set (words)} & 27.4 \\
    \hspace{3mm} - \textit{Chinese Set (chars)} & 35.3 \\
    Avg. Audio Duration (s) & \\ 
    \hspace{3mm} - \textit{English Set} & 10.3 \\
    \hspace{3mm} - \textit{Chinese Set} & 8.5 \\
    \shline
    \end{tabular}
    
    \label{tab:data_statistics}
    \vspace{-2mm}
\end{table}

\subsection{Benchmarks for LALMs}
\label{sec:related_bench}
\paragraph{General Audio Understanding Benchmarks.}
These benchmarks focus on understanding various audio types and reasoning based on audio/speech input. 
Mainstream evaluation forms include speech-based general QAs and instruction following~\cite{voicebench,llama-questions}, or multiple-choice QAs grounded in audio input and textual prompts~\cite{mmau,mmsu,mmar}.
For example, MMAU~\cite{mmau} incorporates a comprehensive coverage of audio domains and task types, focusing on the evaluation of perception and understanding of sounds, speech, and music.
MMAR~\cite{mmar} extends the evaluation to deeper reasoning based on graduate-level and multi-disciplinary knowledge.

\paragraph{Paraliguistic-Aware Spoken Dialogue Benchmarks.}
In response to surging demand for a real-world, empathetic dialogue experience, recent evaluations have increasingly underscored the need for fine-grained perception and response generation involving paralinguistic cues. 
Apart from recognizing semantic meaning, the models are expected to perceive paralinguistic cues implicit in the speaker's voice (such as emotion and age) and generate responses that are appropriate in both speaking styles and content.
Previous work primarily considers a small subset of paralinguistic features to evaluate scenario awareness and speaking style adaptation.
For example, StepEval-Audio-Paralinguistic~\cite{step-audio-2} designs understanding tasks across 11 dimensions and primarily focuses on the Chinese context.
EChat-eval~\cite{echat-eval} encompasses emotion, gender, age, and sound events, with emotional states being the predominant category.
Similarly, ParaS2SBench~\cite{paras2sbench} only includes emotion, sarcasm, gender, and age as paralinguistic cues. 
In contrast, targeting real-world scenarios, our work proposes a comprehensive evaluation suite to assess capabilities in perception and fine-grained generation involving a rich set of paralinguistic features.

\begin{figure}[t!]
    \centering
    \setlength{\abovecaptionskip}{3mm}
    \setlength{\belowcaptionskip}{-6mm}
    \includegraphics[width=0.93\linewidth]{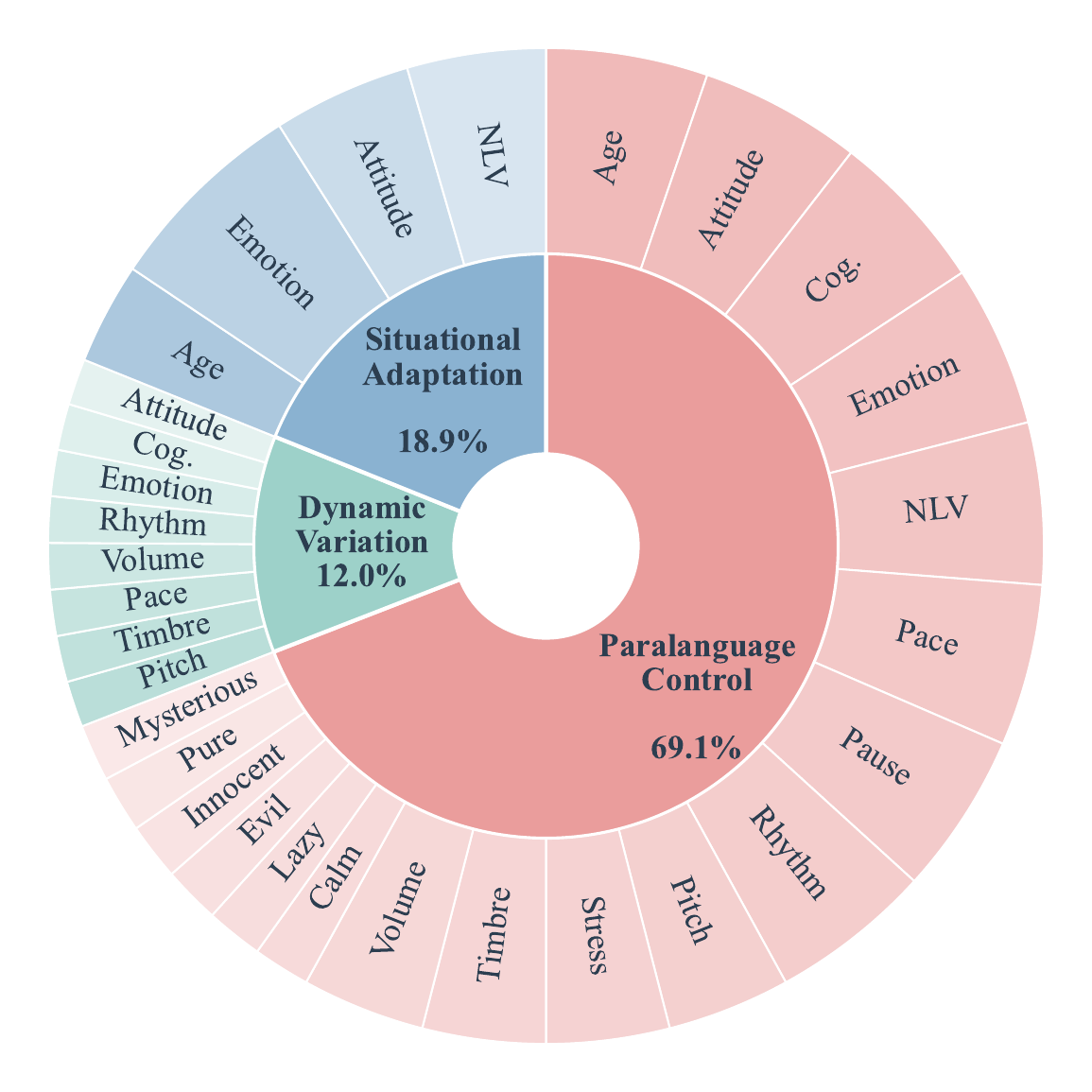} 
    \caption{\textbf{Composition of \rawname.} 
    The dataset consists of over 1,000 bilingual speech queries, encompassing more than 100 paralinguistic features. 
    Cog.: Cognitive State. 
    (See the Appendix for descriptions and value ranges of all dimensions.)}
    \label{fig:overall_composition}
\end{figure}

\subsection{LALM-as-a-Judge for Speech Evaluation}
Due to the high cost and limited scalability of human judgment, LALM-as-a-Judge has become a mainstream evaluation approach for speech-based evaluation.
However, constructing \textit{efficient} and \textit{robust} automatic evaluation pipelines remains a significant open research challenge. 
For instance, S2S-Arena~\cite{s2s-arena} finds that using LALMs as judges for speech evaluation may suffer from severe positional and length biases. Thus, our work relies on manual pairwise comparisons.
To mitigate such positional biases, EmergentTTS-Eval~\cite{EmergentTTS-Eval} introduces randomized ordering for reference-candidate pairs.
Inspired by prior works, we carefully design a suite of prompts and develop a robust evaluation pipeline based on pairwise comparisons. This system significantly reduces reliance on human scoring and enables reliable benchmarking of mainstream LALMs.

\section{\rawname}
\label{sec:bench}

\begin{figure*}[t!]
    \centering
    \includegraphics[width=0.95\linewidth]{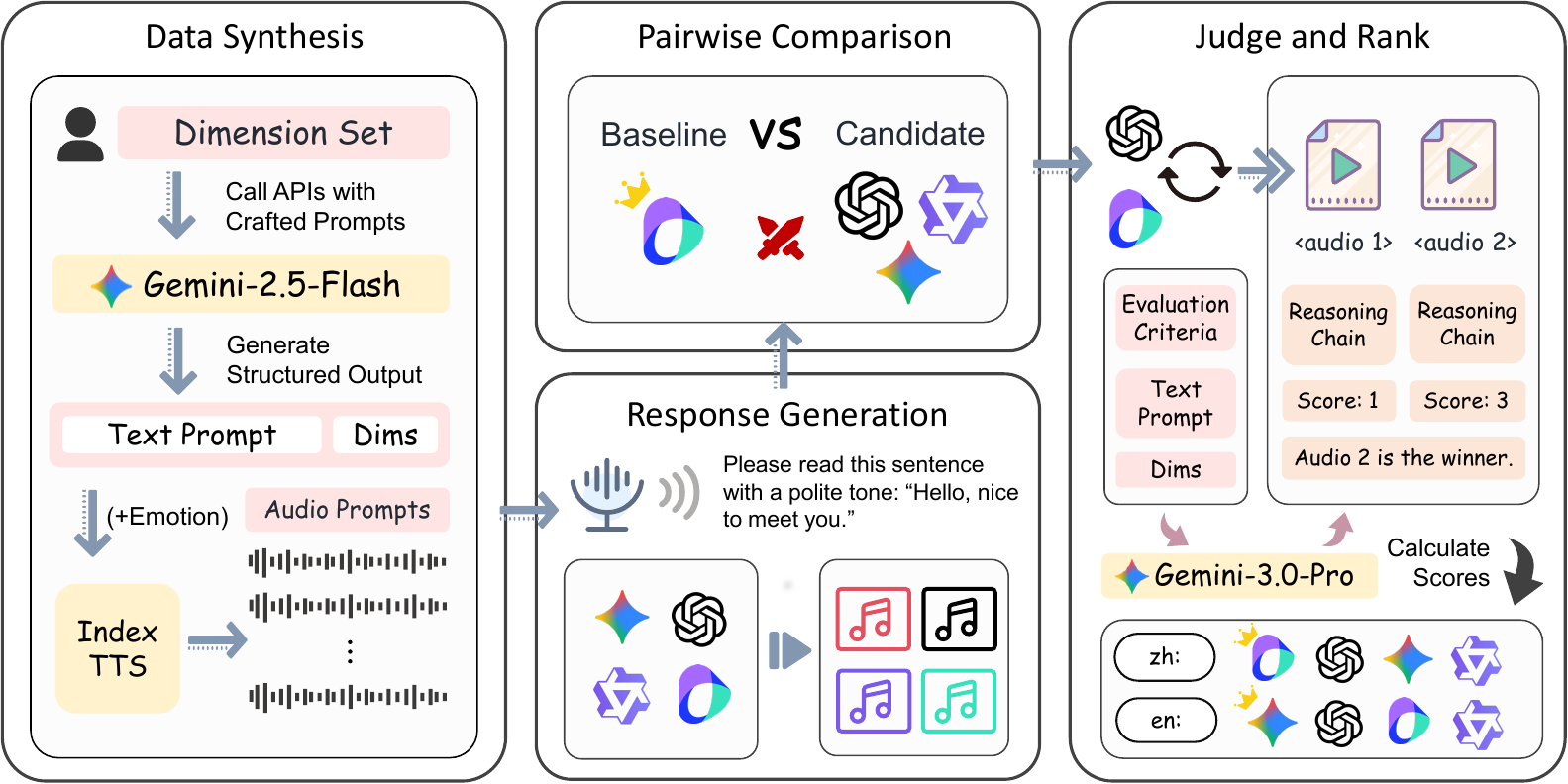}
    
    \caption{\textbf{Overview of the proposed framework.} 
    (1) \textit{Data Engine}: Leverages Gemini to synthesize textual instructions based on a pre-defined dimension set, followed by IndexTTS to generate audio prompts for eliciting LALM responses. 
    (2) \textit{Evaluation Pipeline}: Conducts pairwise comparisons between a strong baseline and candidate models. As the judge, Gemini evaluates audio responses against specific criteria and textual instructions, generating reasoning chains and scores to produce the final leaderboard.
}
    \label{fig:overview}
\end{figure*}

We introduce \name, a comprehensive, Chinese-English parallel benchmark designed to evaluate the capabilities of paralinguistic speech generation. 
The dataset comprises 1,001 samples, each consisting of a speech query paired with specific paralinguistic dimensions. 
The statistics and composition of \rawname are summarized in~\cref{tab:data_statistics} and~\cref{fig:overall_composition}, respectively.

\subsection{Task Design}

Our design principles focus on three aspects: 
(1) covering fine-grained and interpretable paralinguistic dimensions;
(2) designing a reasonable task hierarchy;
and (3) assessing LALMs' paralinguistic understanding and response capabilities in real-world interactive scenarios through contextualized tasks.
Ultimately, the benchmark is structured around critical skill types: \textit{Paralanguage Control}, \textit{Dynamic Variation}, and \textit{Situational Adaptation}.

\paragraph{Paralanguage Control.}
The Paralanguage Control task instructs an LALM to repeat a sentence with specified paralinguistic features.
It directly assesses the model's proficiency in manipulating various vocal characteristics.
This capability is further divided into two sub-categories: control over \textit{common features} and the generation of \textit{abstract styles}. 
The former centers on 12 common paralinguistic features, encompassing dimensions like expressive (\eg, emotion, attitude), prosodic (\eg, pause, stress), and acoustic features (such as timbre and volume).
The latter probes a more holistic and advanced capability, assessing how well the model can understand and render complex speaking styles and tones that require a combination of features, such as generating a ``lively and mischievous voice''.

\paragraph{Dynamic Variation.}
Building on the static capabilities of Paralanguage Control, the Dynamic Variation task evaluates a more advanced skill: the continuous, fine-grained modulation of paralinguistic features within a single utterance.
This task gauges the model's competence in executing smooth and natural-sounding transitions, which are crucial for human-like speech.
We incorporate 8 dimensions for this task, including pitch, speed, and volume, \etc.
Each instruction chooses two distinct values within the same dimension, connected either through transitional relationships (\eg, $\mathbf{Emotion}: \mathsf{Happy} \to \mathsf{Sad}$)
or progressive relationships (\eg, $\mathbf{Volume}: \mathsf{Whisper} \xrightarrow{\textit{increase}} \mathsf{Normal}$).

\paragraph{Situational Adaptation.}
This task emulates real-life empathetic dialogue by incorporating user utterances grounded in specific socio-affective contexts. The model is expected to comprehend the complex scenario and generate responses with appropriate semantic content and speaking style. It evaluates the model's ability to infer paralinguistic cues embedded in speech and produce contextually appropriate spoken utterances. This category mainly involves 4 paralinguistic dimensions, \ie, age, emotion, attitude, and non-linguistic vocalizations (\eg, laughter and sighs).

\begin{table*}[t!]
\centering
\small
\tablestyle{6pt}{1.15}

\caption{\textbf{Comparison with related benchmarks.}
Our benchmark features comprehensive coverage of paralinguistic features, diverse tasks, and an automated, transcription-free evaluation pipeline that directly assesses audio. 
\# Features: total number of paralinguistic features involved.
Para. Con: fine-grained paralanguage control over the speech generation.
Dyn. Var: dynamic variation of the paralinguistic features.
Sit. Ada: situational adaptation in dialogue.
}

\begin{tabular}{lccccccc}
\toprule
\multirow{2}{*}{\textbf{Benchmarks}} &
  \multirow{2}{*}{\textbf{Size}} &
  \multirow{2}{*}{\textbf{\# Features}} &
  \multicolumn{3}{c}{\textbf{Eval. Aspects}} &
  \multirow{2}{*}{\textbf{\begin{tabular}[c]{@{}c@{}}Pairwise\\ Evaluation\end{tabular}}} \\ 
  \cmidrule(lr){4-6}
                              &       &    & \textit{Para. Con} & \textit{Dyn. Var} & \textit{Sit. Ada} &        \\ \midrule

AIR-Bench~\cite{air-bench}                     & 3,000    & 10  & \xmark              & \xmark              & \xmark            & \xmark \\
SD-Eval~\cite{sd-eval}                       & 6,613 & 16  & \xmark              & \xmark              & \cmark            & \xmark \\
S2S-Arena~\cite{s2s-arena}                     & 154   & 11 & \xmark              & \xmark              & \cmark            & \cmark \\
StepEval-Para~\cite{step-audio-2} & 450   & 44  & \xmark              & \xmark              & \xmark            & \xmark \\
TELEval~\cite{televal}                      & 1,540 & 11  & \xmark              & \xmark              & \cmark            & \xmark \\
EChat-eval~\cite{echat-eval}                    & 1400   & 27  & \xmark              & \xmark              & \cmark            & \xmark \\
ParaS2SBench~\cite{paras2sbench}                  & 2,690 & 12  & \xmark              & \xmark              & \cmark            & \xmark \\
VStyle~\cite{vstyle}             & 762 & 16  & \cmark              & \cmark              & \cmark            & \xmark \\
\textbf{\rawname (Ours)} & 1,001 & \textbf{101} & \cmark              & \cmark              & \cmark            & \cmark 
\\

\bottomrule
\end{tabular}
\label{tab:bench_compare}
\end{table*}

\subsection{Data Curation}
To facilitate the construction of evaluation samples, we design an efficient, scalable data pipeline. ~\cref{fig:overview} illustrates our framework, comprising a data engine that synthesizes paralinguistic-related speech queries, and an automated evaluation pipeline that assesses response quality.
We detail the data engine in this section, while the evaluation settings are presented in the subsequent section.

\paragraph{Instruction Synthesis.}
We primarily leverage Gemini 2.5 Flash for instruction synthesis. 
Specifically, we define five common real-world settings (campus, workplace, daily life, family, and entertainment), each accompanied by five representative scenarios. 
The LLM is explicitly instructed to cover these contexts and generate relevant, appropriate queries along with their corresponding paralinguistic dimensions. 
To enhance instruction-following capabilities and data quality, we provide the model with meticulously crafted in-context demonstrations. 
Furthermore, we iteratively input curated dimension sets into the model in batches. This strategy ensures a diverse and balanced distribution across typical real-life scenarios.
This process yields a textual dataset $\mathcal{T} = \{(t_i, \mathtt{Dim}_i)\}_{i=1}^N$, where $t_i$ denotes the textual prompt and $\mathtt{Dim}_i$ represents the associated paralinguistic dimension set (\eg, emotion, age).

For the Paralanguage Control and Dynamic Variation tasks, we adopt a fixed instruction as a prefix, ``\underline{Please read this sentence} \{Dimension(s)\}: \{Sentence\}'', instructing models to repeat a sentence with required paralinguistic dimensions.
For Situational Adaptation, the paralinguistic cues are implicit in the query audio. 
The model needs to derive the contextualized scenarios from user speech and respond with appropriate content and speaking styles.

\begin{table*}[t!]
\tablestyle{13pt}{1.25}
\setlength{\arrayrulewidth}{1.5\arrayrulewidth}
    \centering
    \normalsize

\caption{\textbf{Overall performance comparison of state-of-the-art LALMs on Chinese and English subsets.} 
Models are listed in descending order of overall performance. Style represents abstract style. S-Dim and M-Dim denote single-dimension and multi-dimension, respectively.
The \textbf{first}\,\protect\textcolor{gold}{\faMedal}, \underline{second}\,\protect\textcolor{silver}{\faMedal}, and third\,\protect\textcolor{bronze}{\faMedal} places of each evaluation module are highlighted.}

\resizebox{.98\linewidth}{!}{%
\begin{tabular}{lccccccccc}
\toprule
\multirow{2}{*}{\textbf{Model}} &
  \multirow{2}{*}{\textbf{Overall}} &
  \multicolumn{4}{c}{\textbf{Paralanguage Control}} &
  \multirow{2}{*}{\textbf{\thead{Dynamic\\Variation}}} &
  \multicolumn{3}{c}{\textbf{Situational Adaptation}} \\
  \cmidrule(rl){3-6} \cmidrule(rl){8-10}
 &
 &
  \textit{Style} &
  \textit{S-Dim} &
  \textit{M-Dim} &
  \textit{Total} &
  &
  \textit{S-Dim} &
  \textit{M-Dim} &
  \textit{Total} \\
   \midrule
\rowcolor{gray!30}
\textsc{Chinese (zh)} &&&&&&&&&\\
Doubao Realtime Voice  & \best{70.84} & 80.03 & 71.77 & 68.87 & \best{71.86} & \second{54.09} & 54.39 & 58.10 & \best{58.21}  \\
GPT Audio              & \second{39.09} & 24.77 & 37.50 & 36.58 & \second{35.57} & \best{63.33} & 43.50 & 38.33 & 40.18  \\
Gemini Audio & \third{28.18} & 20.72 & 26.19 & 36.05 & \third{29.64} & 29.17 & 29.00 & 19.72 & 23.04 \\
Qwen3-Omni-Flash       & 22.58 & 7.66  & 14.29 & 15.92 & 14.16 & \third{35.00} & 46.50 & 43.61 & \third{44.64} \\
Qwen3-Omni-Realtime    & 14.34 & 2.25  & 4.76  & 5.39  & 4.72 & 5.83 & 51.50 & 48.06 & \second{49.29} \\

\midrule[0.05pt]
\midrule[0.05pt]

\rowcolor{gray!30}
\textsc{English (en)} &&&&&&&&&\\
Gemini Audio & \best{64.97} & 65.09 & 68.43 & 62.80 & \best{66.49} & \best{61.08} & 52.01 & 51.21 & \second{52.37} \\
GPT Audio              & \second{49.39} & 45.95 & 43.69 & 49.47 & \second{46.38} & \second{52.92} & 58.00 & 57.50 & \best{57.68} \\
Doubao Realtime Voice  & \third{31.39} & 25.68 & 26.19 & 30.79 & \third{28.05} & \third{22.50} & 46.00 & 46.11 & 46.07  \\
Qwen3-Omni-Realtime    & 15.52 & 0.90  & 7.74  & 7.37  & 6.75 & 5.00 & 44.00 & 51.11 & \third{48.57}  \\
Qwen3-Omni-Flash       & 13.73 & 9.46  & 10.71 & 15.39 & 12.51  & 7.92 & 22.00 & 19.17 & 20.18 \\

\bottomrule
\end{tabular}
}
\label{tab:overall-compare}
\end{table*}

\paragraph{Speech Synthesis.}
To convert the previously acquired textual instructions into speech queries, we utilize a robust open-source Text-to-Speech (TTS) model~\cite{indextts2} known for its zero-shot timbre reconstruction capabilities and fine-grained control over emotional tones.
The synthesis process with the TTS-based system can be formulated as:
\begin{equation}
    s_i = \text{TTS}(t_i, \mathtt{Dim}_i \mid \mathbf{a}_{ref})
    \text{,}
\end{equation}
where $\mathbf{a}_{ref}$ denotes the reference audio used for timbre control.
For the \textit{Paralanguage Control} and \textit{Dynamic Variation} tasks, we use a fixed male-voice reference audio with a neutral tone.
For the \textit{Situational Adaptation} task, we craft a delicate scheme to align style with content.
Specifically, the age and attitude dimensions are controlled by the timbre and style prompts (reference audio clips), while the emotion dimension is modulated by the emotion vector.
We note that non-linguistic vocalizations can be seamlessly integrated into the prompts via textual hints (\eg, ``Ah'', ``Cough'') without requiring special treatment.
Finally, we obtain the multimodal evaluation dataset $\mathcal{D}_{eval} = \{(s_i, t_i, \mathtt{Dim}_i)\}_{i=1}^N$.

\paragraph{Quality Check.}
We conduct a rigorous manual check of the constructed samples and ensure that the data quality meets the criteria.  
Critical aspects include whether the synthesized speech $s_i$ is clear and recognizable, and whether the speaking content $t_i$ and the corresponding paralinguistic dimensions $\mathtt{Dim}_i$ are reasonable in real-life scenarios.

\subsection{Pairwise Evaluation Pipeline}

\paragraph{General Setting.}
In this work, we employ Gemini 3 Pro~\cite{gemini-3} as an LALM-based judge, owing to its superior capabilities in audio perception and reasoning.
Following the protocol in~\cite{EmergentTTS-Eval}, we devise a baseline-candidate evaluation framework based on pairwise comparisons.
In this setup, a set of candidate models $\mathcal{M} = \{M_k\}_{k=1}^K$ are evaluated against a fixed baseline $B$.
Specifically, given a speech query $s_i$, the baseline model $B$ and the candidate model $M_k$ generate their speech responses, denoted as $r_i^{B}$ and $r_i^{M_k}$, respectively.
Given the response pair, the query transcript $t_i$, the target dimensions $\mathtt{Dim}_i$, and the evaluation criteria $\mathcal{C}_{eval}$, the LALM judges which response is better:
\begin{equation}
    w_i=\mathcal{J}\left(r_i^{B}, r_i^{M_k}, t_i, \mathtt{Dim}_i \mid \mathcal{C}_{eval}\right)
    \text{,}
\end{equation}
where $w_i\in\{0,1,2\}$ denotes the winner index, and 0 denotes a tie. 
For each sample, only the winner receives 1 score; in the event of a tie, both models receive 0.5 points.
In practice, we use Doubao Realtime Voice Model and Gemini Audio as Chinese and English baselines, respectively.

\paragraph{Bias and Hallucination Control.}
To mitigate the judgment bias brought by response orders, we follow EmergentTTS-Eval~\cite{EmergentTTS-Eval} to randomly assign orders for the baseline and candidate model.
For judgment robustness and accuracy, we prompt the judge with a Chain-of-Thought (CoT) strategy. The judge is explicitly required to analyze specific aspects before rating (on a 0-3 Likert scale) and selecting the winner.
For Paralanguage Control and Dynamic Variation tasks, the evaluated aspects include \textit{Content Accuracy}, \textit{Fluency and Naturalness}, and \textit{Paralinguistic Compliance}.
For the Situational Adaptation task, the corresponding judging aspects include \textit{Content Relevance}, \textit{Fluency and Naturalness}, and \textit{Paralinguistic Alignment}.

To reduce hallucinations, the LALM judge is explicitly prompted to ground analysis in specific timestamps from the corresponding audio clips (a detailed evaluation prompt is available in the Appendix).

\paragraph{Judging Metrics.}
Model performance is evaluated by aggregating scores either task-wise or across the entire set.
To mitigate potential bias arising from the varying capabilities of candidate models, we introduce a weighted scoring mechanism for the baseline performance.
Let $S(M_k, \tau)$ denote the total score of candidate $M_k$ on task $\tau$.
The performance of the baseline is defined as the weighted average of its pairwise scores against all candidates:
\begin{equation}
S(B, \tau) = \sum_{k=1}^{K} \left( S(B, \tau \mid M_k) \times \frac{S(M_k, \tau)}{\sum_j S(M_j, \tau)} \right)
\text{,}
\end{equation}
where $S(B, \tau \mid M_k)$ is the score of the baseline against the candidate.
The core intuition behind this weighting scheme is to adjust the contribution of each pairwise comparison based on the opponent's relative strength~\cite{chatbot-arena}. 
Unless otherwise specified, all reported results are normalized to a range of 0-100\% to facilitate comparisons within each task.

\subsection{Comparison with Existing Benchmarks}
As summarized in~\cref{tab:bench_compare}, \name distinguishes itself from existing benchmarks in three key aspects: 
(1) Broad Paralinguistic Coverage: We cover a wider range of paralinguistic features derived from common real-life interactions.
(2) Application-Oriented Task Hierarchy: We design a progressive task structure, ranging from basic paralanguage control (useful in applications such as role-play) to continuous fine-grained modulation (\eg, for storytelling and news reports), and finally to social-affective understanding and response generation (for scenarios like empathetic dialogue and social companionship).
(3) Scalable Speech-based Evaluation: We introduce an LALM-based evaluation pipeline. Unlike prior methods that rely on distortion-prone transcriptions or expensive human labor, our approach enables robust, efficient speech assessment.

\section{Empirical Results and Analysis}
\label{sec:exp_res}

Our main objective is to evaluate the most advanced LALMs on generation capabilities with paralinguistic features.
The models include gpt-audio-2025-08-28~\cite{gpt0828}, Gemini 2.5 Flash Audio~\cite{gemini-audio}, Doubao Realtime Voice Model~\cite{doubao-voice}, Qwen3-Omni-Flash-2025-12-01~\cite{qwen3_omni_20251201}, and Qwen3-Omni-Realtime~\cite{qwen3-omni}, accessed via APIs using default decoding parameters.
 We evaluate these models on both English and Chinese subsets of our benchmark.

\subsection{Main Results}

\paragraph{Paralanguage Control.}
As shown in~\cref{tab:overall-compare}, Doubao holds a commanding lead on the Chinese domain (71.86). For the English benchmark, however, Gemini (66.49) takes the lead, GPT performs more consistently across both the Chinese (35.57) and English (46.38) sections.

\begin{figure}[!htbp]
    \centering
    \setlength{\abovecaptionskip}{-1mm}
    \setlength{\belowcaptionskip}{-6mm}
    \includegraphics[width=0.9\linewidth]{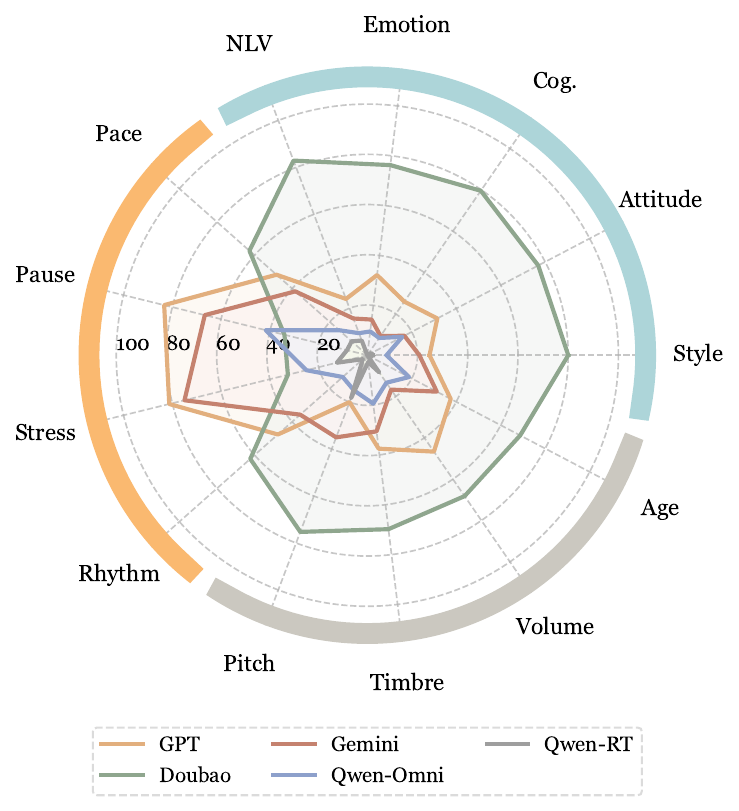} 
    \caption{\textbf{Dimension-wise performance on Paralanguage Control task (zh)}.
    We categorize paralinguistic dimensions into \colorbox{expressive_color}{Expressive}~(NLV, Emotion, Cog., Attitude, Style), \colorbox{prosodic_color}{Prosodic}~(Pace, Pause, Stress, Rhythm), and \colorbox{acoustic_color}{Acoustic}~(Pitch, Timbre, Volume, Age) features.}
    \label{fig:task-perf}
\end{figure}

As shown in~\cref{fig:task-perf}, different models excel in different paralinguistic dimensions.
Doubao is relatively more well-rounded. It performs better on expressive features and acoustic features, but still lags in prosodic attributes. 
In contrast, GPT and Gemini significantly outperform in pause and stress, but lag in expressive dimensions.

The models exhibit significant regional variations in their core capabilities, primarily resulting from differences in training corpora and inherent language characteristics. While Chinese-centric models place greater emphasis on localized expressions and acoustic nuances, English-centric models perform better in terms of prosodic structure.

\begin{takeaway}
Achieving comprehensive static control still poses a significant challenge for LALMs.
\end{takeaway}

\paragraph{Dynamic Variation.}
As shown in \cref{tab:average_score}, Dynamic Variation constitutes the primary bottleneck for current LALMs, achieving the lowest average score (56.51/100) among all tasks. This difficulty likely stems from the strong coupling between paralinguistic features and linguistic content, as well as the scarcity of training data exhibiting explicit intra-utterance variation, which hinders models from learning fine-grained and controllable modulation.

Notably, model behaviors differ significantly. GPT shows limited but observable dynamic adjustment, whereas the Qwen3 series models exhibit weaker instruction-following ability in this task, often failing to translate variation instructions into appropriate acoustic changes.

\begin{takeaway}
Basic dynamic modulation of paralinguistic features continues to represent a widespread bottleneck.
\end{takeaway}

\begin{table}[t!]
    \centering
    \small
    \tablestyle{12pt}{1.1}

    \caption{\textbf{Initial average score assigned by judge model.} 
    The average of the initial scores of all evaluated models given by the judge model~(scores normalized to 0–100).} 
    
\begin{tabular}{l r} 
    \shline
    \textbf{Task} & {\textbf{Average Score}} \\
    \hline
    Situational Adaptation & 68.64 \\ 
    Paralanguage Control   & 66.01 \\ 
    Dynamic Variation      & 56.51 \\ 
    \shline
    \end{tabular}
    
    \label{tab:average_score}
    \vspace{-4.5mm}
\end{table}

\paragraph{Situational Adaptation.}
Human judgment reveals that models exhibit significant shortcomings when handling complex human interaction logic, as discussed in~\cref{para: Failure Analysis}.

While leading LALMs can generally understand and respond well to simple paralinguistic features in user instructions, they sometimes struggle to grasp more complex paralinguistic features (\eg, sarcastic tone), indicating that LALMs still lack the ability to analyze and capture complex paralinguistic information within a speech segment. Limited performance in role adaptation may be related to the ``voice assistant'' identity constraints reinforced during the pre-training phase, making it difficult for the model to fully immerse itself in other complex emotional contexts.

\begin{takeaway}
It remains a challenge for LALMs to grasp and respond to contextualized or abstract paralinguistic information in user's speech.
\end{takeaway}

\definecolor{tagbg}{RGB}{235, 242, 250}
\definecolor{tagtext}{RGB}{44, 62, 80}
\definecolor{analysisgray}{RGB}{245, 245, 245}

\newcommand{\paradim}[2]{%
  \setlength{\fboxsep}{1.5pt}%
  \raisebox{-2pt}{%
  \adjustbox{valign=m}{%
    \colorbox{tagbg}{%
      \footnotesize\sffamily\textbf{\textcolor{tagtext}{#1}}: \textcolor{tagtext}{#2}%
        }%
    }%
  }%
}

\newcommand{\turn}[2]{\textbf{\sffamily #1:} #2}

\begin{table*}[t!]
\centering
\small

\caption{\textbf{Manual qualitative analysis of failure modes} of Gemini Audio.
We show representative cases of each type with detailed human-annotated analysis.
For ease of visualization, all cases are translated into English. Para. Dim(s): paralinguistic dimension(s).
}\label{tab:case_study}

\begin{adjustbox}{max width=0.98\linewidth}
\begin{tabularx}{\linewidth}{l X}
\toprule
\textbf{Text Prompt} & Please speak to the user in an appropriate tone based on the user's pre-set scenario. \\ 
\midrule
\textbf{Failure Mode} & \textbf{Case Study} \\
\midrule

Role-play Failure & 
\turn{Input Audio}
\textbf{Context:} Haha, long time no see, come in and sit, I've missed you so much! \par
\vspace{1pt}
\hspace{6.0em} \textbf{Para. Dim(s):} \paradim{Attitude}{Enthusiastic} \paradim{NLV}{With Laughter} \par
\vspace{3pt}
\turn{Gemini Output Audio}{It sounds like you two are very close. I'm glad you've reunited.} \par
\vspace{3pt}
\colorbox{analysisgray}{\parbox{\dimexpr\hsize-2\fboxsep\relax}{%
  \emph{\textbf{Analysis:} The model misinterprets the subject ``you'', incorrectly assuming the user is talking to another person and thus responding as a third-party observer.}%
}}
\\ 
\midrule

Neglecting Paralanguage & 
\turn{Input Audio}
\textbf{Context:} Look at what you've done, truly ``perfect'' isn't it? \par
\vspace{1pt}
\hspace{6.0em} \textbf{Para. Dim(s):}\paradim{Attitude}{Sarcastic} \par
\vspace{2pt}
\turn{Gemini Output Audio}{I'm not sure what you're referring to, but I'm glad you think it's good! Would you like to chat about your thoughts?} \par
\vspace{3pt}
\colorbox{analysisgray}{\parbox{\dimexpr\hsize-2\fboxsep\relax}{%
  \emph{\textbf{Analysis:} The model fails to detect the user's sarcastic attitude. It interprets the input literally as praise, and consequently provides an inappropriately enthusiastic response.}%
}}
\\ 
\midrule

Template Response & 
\turn{Input Audio}
\textbf{Context:} Honey, your little cat drawing is so cute! Haha! \par
\vspace{1pt}
\hspace{6.0em} \textbf{Para. Dim(s):}\paradim{NLV}{With Laughter} \par
\vspace{2pt}
\turn{Gemini Output Audio}{The children's paintings are full of childlike innocence and creativity.} \par
\vspace{3pt}
\colorbox{analysisgray}{\parbox{\dimexpr\hsize-2\fboxsep\relax}{%
  \emph{\textbf{Analysis:} The model overlooks the user's intimate tone and the specific 'cat drawing' detail, instead defaulting to a generic, formalized comment on children's art.}%
}}
\\ 
\midrule

Intent Misunderstanding & 
\turn{Input Audio}
\textbf{Context:} Our class monitor's speech was so funny, haha, my stomach hurts from laughing. \par
\vspace{1pt}
\hspace{5.9em} \textbf{Para. Dim(s):}\paradim{NLV}{With Laughter} \par
\vspace{2pt}
\turn{Gemini Output Audio}{A good speech can leave a deep impression.} \par
\vspace{3pt}
\colorbox{analysisgray}{\parbox{\dimexpr\hsize-2\fboxsep\relax}{%
  \emph{\textbf{Analysis:} The model overlooks the user's humor and the paralinguistic ``laughter'' cue. It adopts an inappropriately formal tone and responds to a joke with a flat attitude.}%
}}
\\
\bottomrule
\end{tabularx}
\end{adjustbox}
\end{table*}

\subsection{Agreement with Human Judgment}
\paragraph{Setup.}
Following the previous approach~\cite{EmergentTTS-Eval}, we conduct a human evaluation to test how well model-based judgment aligns with human subjective preference.
Human judges are instructed to determine the winner (or tie) of each response pair, with instructions and evaluation criteria similar to those in the model-judging prompt.
We randomly select 5\% samples from each task, totaling 416 response audio pairs.

\paragraph{Correlation Analysis.}
We compute Spearman's rank correlation coefficient to measure the agreement between rankings derived from model judgments and those from human evaluators.
Our automated speech-based evaluation system achieves correlation scores of 0.90 and 1.00 on the Chinese and English subsets, respectively, both of which are statistically significant.
These results indicate that our evaluation pipeline closely aligns with human preferences when assessing audio pairs with paralinguistic features.

\begin{figure}[!htp]
    \centering
    \setlength{\abovecaptionskip}{0mm}
    \setlength{\belowcaptionskip}{-4.5mm}
    \includegraphics[width=0.78\linewidth]{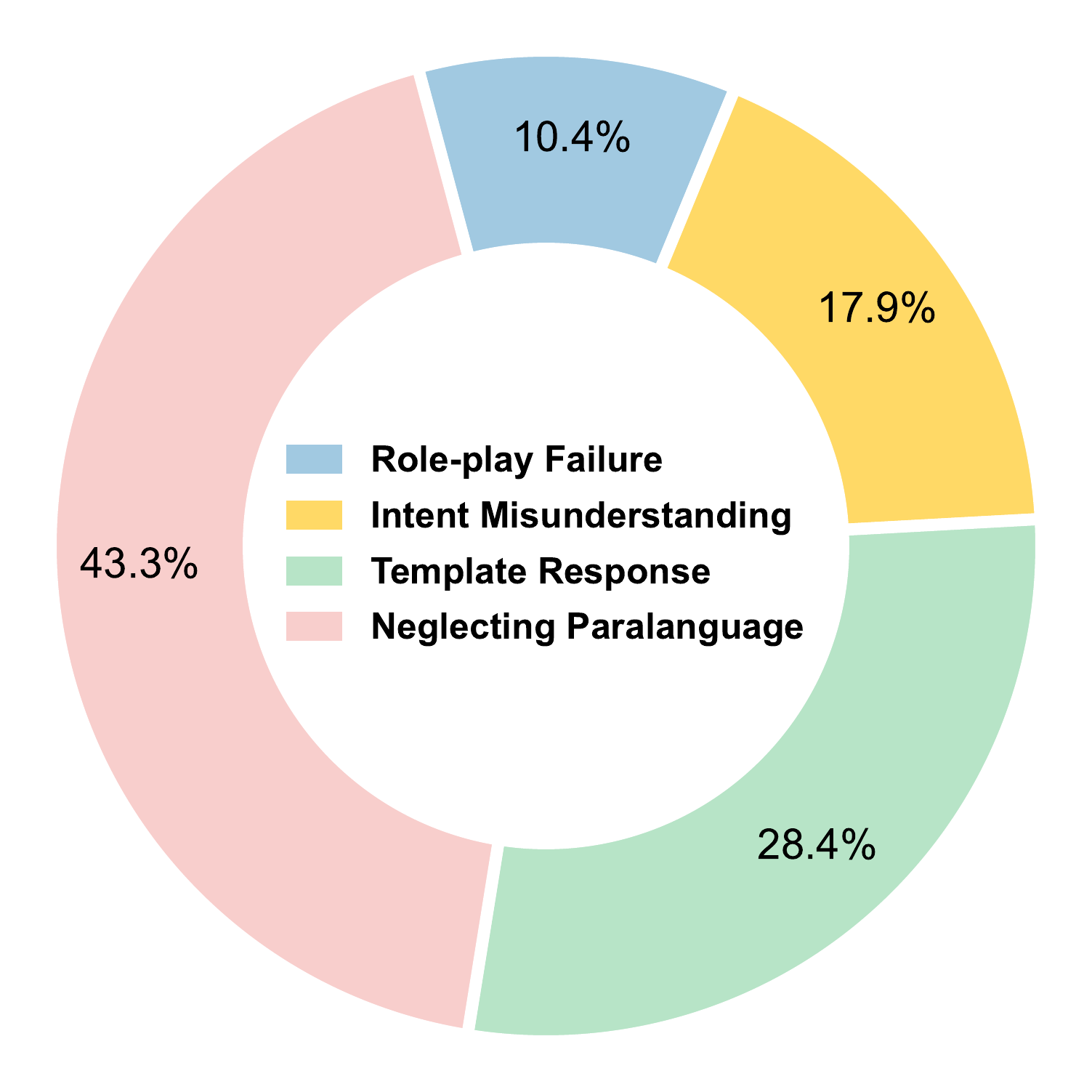} 
    
    \caption{\textbf{Distribution of error types} of Gemini Audio on the Situational Adaptation task. }
    \label{fig:error_analysis}
\end{figure}

\subsection{Failure Analysis} \label{para: Failure Analysis}
We conduct a manual failure analysis for Gemini Audio on the Situational Adaptation task in the Chinese subset to identify potential failure modes.
In total, Gemini Audio fails in 67 out of 190 samples, which can be categorized into four types: \textit{Role-play Failure}, \textit{Intent Misunderstanding}, \textit{Template Response}, and \textit{Neglecting Paralanguage}.
As illustrated in~\cref{fig:error_analysis}, a substantial portion (43.3\%) of these errors stems from overlooking paralinguistic information embedded in the user's speech.
These results underscore the importance of understanding paralanguage alongside linguistic content for enhanced human-computer interaction.
We present representative cases and detailed analyses of each pattern in~\cref{tab:case_study}.
\section{Conclusion}
We introduce~\rawname, a comprehensive benchmark for paralinguistic-aware speech generation.
Designed for real-world scenarios, it features three tasks of increasing complexity: Paralanguage Control, Dynamic Variation, and Situational Adaptation.
Using our curated automatic evaluation pipeline, we reveal limitations in leading voice assistants' ability to generate natural speech with nuanced paralanguage.
Our findings highlight potential avenues for improving LALMs and underscore the need for models with enhanced paralinguistic capabilities.

{\small
\bibliographystyle{ieeenat_fullname}
\bibliography{reference}
}

\ifarxiv \clearpage \appendix \onecolumn \section{Descriptions of Paralinguistic Features}
We list paralinguistic dimensions, their descriptions, and possible values (\ie, paralinguistic features) below.
Our benchmark includes more than 100 paralinguistic features across 13 dimensions.

\begin{table}[!htbp]
    \label{tab:paralinguistic_features}
    \begin{tabularx}{\textwidth}{@{} l >{\raggedright\arraybackslash}X >{\raggedright\arraybackslash}X @{}}
        \toprule
        \textbf{Paralinguistic Dimension} & \textbf{Dimension Description} & \textbf{Paralinguistic Feature} \\
        \midrule
        Age & Refers to the speaker's age group & child, youthful, adult, elderly \\
        \addlinespace
        Pitch & Refers to the frequency of the speaker's voice & very high pitch, high pitch, medium pitch, low pitch, very low pitch \\
        \addlinespace
        Timbre & Refers to the qualitative characteristics of the speaker's voice & bright, hoarse, smooth, rich, gentle, sweet \\
        \addlinespace
        Pace & Refers to the speed of speech & very fast pace, fast pace, medium pace, slow pace, very slow pace \\
        \addlinespace
        Volume & Refers to the loudness of the speaker's voice & shouting manner, loudly, normal volume, quietly, whisper \\
        \addlinespace
        Pause & Refers to interruptions during speech & with a clear pause after the a specific word \\
        \addlinespace
        Rhythm & Refers to the regular variation in speech & steady rhythm, lighthearted rhythm, soothing rhythm, rushed rhythm, emphatic rhythm, dragging rhythm, halting rhythm \\
        \addlinespace
        Stress & Refers to the emphasis placed on words during speech & with emphasis on, with stress on, with heavy stress on, with a forceful tone on \\
        \addlinespace
        Emotion & Refers to the feelings expressed during speech & neutral emotion, happy emotion, sad emotion, angry emotion, surprised emotion, disgusted emotion, fearful emotion \\
        \addlinespace
        Cognitive State & Refers to the speaker's state of mind during the speech process & confident tone, hesitant tone, confused tone, doubting tone, tired tone, curious tone, anxious tone, helpless tone, nervous tone \\
        \addlinespace
        Non-Linguistic Vocalizations & Refers to sounds made during speech that do not carry semantic meaning & with laughter, with crying, with a sigh, with coughing, with a scream, with hiccups, with a yawn, with a smack of the lips \\
        \addlinespace
        Attitude & Refers to the speaker's subjective stance toward the listener & polite tone, sincere tone, enthusiastic tone, cold tone, sarcastic tone, contemptuous tone, rude tone, perfunctory tone, teasing tone \\        
        \addlinespace
        Style & Refers to the distinctive manner or character persona adopted during speech & with a lively and mischievous voice, with a professional and objective tone, with an evil tone, with a lazy and casual manner, with a mysterious and unpredictable tone, with a serious and earnest tone, with an innocent and pure voice ... \\
        \bottomrule
    \end{tabularx}
\end{table}

\section{Prompt Templates}
We present prompt templates used in our data pipeline, including those for LLM-based instruction synthesis and LALM-based evaluation.
\subsection{Instruction Synthesis Prompt}
In our data synthesis pipeline, we prompt Gemini 2.5 Flash to synthesize desired textual prompts and their corresponding paralinguistic dimensions, producing a structured output. The prompt is shown below.
\begin{center}
\begin{tcolorbox}[colback=gray!00,
                  colframe=black,
                  width=17.2cm,
                  arc=1.5mm, auto outer arc,
                  breakable,
                  left=0.9mm, right=0.9mm,
                  boxrule=0.9pt,
                  title = {Query prompt of LLM in instruction synthesis}
                 ]
You are a senior expert in speech generation and paralinguistics, proficient in designing test instructions to accurately evaluate a speech model's paralinguistic generation capabilities, and skilled in both Chinese and English. Your task is to create high-quality, diverse, and challenging Chinese and English test instructions for a project evaluating the paralinguistic generation of a speech model.

\vspace{2mm}
\noindent \textbf{[The Start of User Instruction]}
\{ \\
\textbf{Task Description:} 
\begin{adjustwidth}{2em}{0em}
Based on the following classification and collection of paralinguistic features, generate unique, high-quality test instructions in JSONL format. Each instruction should include a prompt containing content that the user requests the speech model to recite, as well as a structured context object (dimensions) that precisely describes the paralinguistic features required in the generated speech, aiming to assess whether the speech model can correctly produce content according to the specified paralinguistic feature requirements.
\end{adjustwidth}
\vspace{2mm}
\textbf{Paralinguistic Features:} 
\begin{adjustwidth}{2em}{0em}
$\bullet$ Age: Refers to the speaker's age group, including child, youthful, adult, elderly.

$\bullet$ Pitch: Refers to the frequency of the speaker's voice, including very high pitch, high pitch, medium pitch, low pitch, very low pitch.

$\bullet$ Timbre: Refers to the qualitative characteristics of the speaker's voice, including bright, hoarse, smooth, rich, gentle, sweet.

$\bullet$ Pace: Refers to the speed of speech, including very fast pace, fast pace, medium pace, slow pace, very slow pace.

$\bullet$ Volume: Refers to the loudness of the speaker's voice, including shouting manner, loudly, normal volume, quietly, whisper.

$\bullet$ Pause: Refers to interruptions during speech, including with a clear pause after a specific word

$\bullet$ Rhythm: Refers to the regular variation in speech, including steady rhythm, lighthearted rhythm, soothing rhythm, rushed rhythm, emphatic rhythm, dragging rhythm, halting rhythm.

$\bullet$ Stress: Refers to the emphasis placed on words during speech, including with emphasis on, with stress on, with heavy stress on, with a forceful tone on.

$\bullet$ Emotion: Refers to the feelings expressed during speech, including neutral emotion, happy emotion, sad emotion, angry emotion, surprised emotion, disgusted emotion, fearful emotion.

$\bullet$ Cognitive State: Refers to the speaker's state of mind during the speech process, including confident tone, hesitant tone, confused tone, doubting tone, tired tone, curious tone, anxious tone, helpless tone, nervous tone.

$\bullet$ Non-Linguistic Vocalizations: Refers to sounds made during speech that do not carry semantic meaning, including with laughter, with crying, with a sigh, with coughing, with a scream, with hiccups, with a yawn, with a smack of the lips.

$\bullet$ Attitude: Refers to the speaker's subjective stance toward the listener, including polite tone, sincere tone, enthusiastic tone, cold tone, sarcastic tone, contemptuous tone, rude tone, perfunctory tone, teasing tone.

Set of paralinguistic feature dimensions: \{SET\_PLACEHOLDER\}
\end{adjustwidth}

\vspace{2mm}
\textbf{Generation Requirements:}
\begin{adjustwidth}{2em}{0em}
1. Generate 3 sets of unique Chinese and English instructions for each element in the set. Each set must meet the generation requirements, have the exact same meaning (differing only in language), and include only one element from the set as the paralinguistic feature requirement.\\
2. The sentences requested for repetition must match the required paralinguistic features to form a natural and credible expression.\\
3. The sentences for the voice model to repeat should be approximately 50 characters/words long, without hint markers such as () or [].\\
4. Scenarios should cover daily life scenes:
\begin{adjustwidth}{2em}{0em}
$\bullet$ Daily routines (washing up, dressing/grooming, making the bed, morning skincare, bedtime relaxation)\\
$\bullet$ Campus (dining in the canteen, self-study in the library, dormitory chatting, classroom interaction, club activities)\\
$\bullet$ Workplace (office collaboration, meeting discussions, remote work, professional socializing, project reporting)\\
$\bullet$ Family (parent-child dialogue, dinner table conversation, family gatherings, doing housework together, bedtime stories)\\
$\bullet$ Entertainment (video games, outdoor travel, watching movies, friend gatherings, sports and fitness)
\end{adjustwidth}
5. Ensure the text content is rich and diverse, avoiding repetitive scenarios and expressions.
\end{adjustwidth}
\vspace{2mm}
\textbf{Generation Output Format:}

\begin{adjustwidth}{2em}{0em}
$\bullet$ JSONL format, one JSON object per line. Each set of instructions occupies two lines without breaks between sets: one line in Chinese and one line in English, identical in meaning, differing only in language, and both adhering to the corresponding requirements.\\
$\bullet$ dimensions must correspond to the paralinguistic feature dimension of the element in the set.\\
{\begin{CJK}{UTF8}{gbsn}
$\bullet$ 中文：\{\textquotedbl prompt\textquotedbl:\textquotedbl 请用[集合中元素]的[对应副语言特征]说:[让语音模型复述的内容]\textquotedbl, \textquotedbl dimensions\textquotedbl: [\textquotedbl 考察的副语言特征\textquotedbl ]\}。注意 prompt 部分请组织成自然的语言表达形式。\\
\end{CJK}}
$\bullet$ English: \{\textquotedbl prompt\textquotedbl :\textquotedbl Please read this sentence with a [elements in the set] voice(or any other suitable expression):\textquotesingle [the content for the speech model to repeat]\textquotedbl \textquotesingle , \textquotedbl dimensions\textquotedbl : [\textquotedbl the examined paralinguistic features\textquotedbl ]\}. Please make sure the prompt is phrased in natural language.
\end{adjustwidth}
\vspace{2mm}
\textbf{Output Example:}
\begin{adjustwidth}{2em}{0em}
{\begin{CJK}{UTF8}{gbsn}
\{\textquotedbl prompt\textquotedbl : \textquotedbl 请用孩童的声音说：妈妈，我可以再吃一块饼干吗？\textquotedbl , \textquotedbl 
dimensions\textquotedbl : [\textquotedbl 年龄\textquotedbl ]\}\\
\end{CJK}}
\end{adjustwidth}
\begin{adjustwidth}{2em}{0em}
\vspace{-\baselineskip}
\{\textquotedbl prompt\textquotedbl : \textquotedbl Please read this sentence with a child's voice: \textquotesingle Mom, can I have another cookie?\textquotesingle \textquotedbl , \textquotedbl dimensions\textquotedbl : [\textquotedbl Age\textquotedbl ]\}\\
\end{adjustwidth}
\begin{adjustwidth}{2em}{0em}
\vspace{-\baselineskip}
{\begin{CJK}{UTF8}{gbsn}
\{\textquotedbl prompt\textquotedbl : \textquotedbl 请用高音说：早上好！今天天气真不错，心情很好。\textquotedbl , \textquotedbl dimensions\textquotedbl : [\textquotedbl 音高\textquotedbl ]\}\\
\end{CJK}}
\end{adjustwidth}
\begin{adjustwidth}{2em}{0em}
\vspace{-\baselineskip}
\{\textquotedbl prompt\textquotedbl : \textquotedbl Please read this sentence with a high pitch: \textquotesingle Good morning! The weather is really nice today, I feel great.\textquotesingle \textquotedbl , \textquotedbl dimensions\textquotedbl : [\textquotedbl Pitch\textquotedbl ]\}\\
\end{adjustwidth}
\begin{adjustwidth}{2em}{0em}
\vspace{-\baselineskip}
{\begin{CJK}{UTF8}{gbsn}
\{\textquotedbl prompt\textquotedbl : \textquotedbl 请用圆润的音质说：您好，请这边坐，为您准备了茶点。\textquotedbl , \textquotedbl dimensions\textquotedbl : [\textquotedbl 音质\textquotedbl ]\}\\
\end{CJK}}
\end{adjustwidth}
\begin{adjustwidth}{2em}{0em}
\vspace{-\baselineskip}
\{\textquotedbl prompt\textquotedbl : \textquotedbl Please read this sentence with a smooth timbre: \textquotesingle Hello, please have a seat here, I've prepared refreshments for you.\textquotesingle \textquotedbl , \textquotedbl dimensions\textquotedbl : [\textquotedbl Timbre\textquotedbl ]\}\\
……
\}
\end{adjustwidth}
\noindent \textbf{[The End of User Instruction]}

Now please strictly follow the generation requirements, think carefully, and generate the required number of instruction samples based on the above requirements. Please output the results directly.

\end{tcolorbox}
\end{center}

\subsection{Evaluation Prompt}
In our automated speech-based evaluation system, we prompt Gemini 3.0 Pro to judge the winner (or a tie) among an audio response pair. 
In addition to audio responses, we provide textual prompts, corresponding paralinguistic dimensions, and evaluation criteria as inputs to the model.
The prompt is shown below.

\begin{center}
\begin{tcolorbox}[colback=gray!00,
                  colframe=black,
                  width=17.2cm,
                  arc=1.5mm, auto outer arc,
                  breakable,
                  left=0.9mm, right=0.9mm,
                  boxrule=0.9pt,
                  title = {Template prompt of scoring evaluation}
                 ]
\textbf{(System Prompt)} \\
You are a professional LALM audio generation quality evaluator (Judger). Please strictly and fairly evaluate the provided audio file based on the following scoring criteria.
Your goal is to judge two audios generated by large language models according to user's command and analyze which audio demonstrates superior quality across the paralinguistic feature. 
You will receive an audio file **T1** first, and after my prompt, you will receive the second audio file **T2**.\\
\textbf{(Instruction)} \\
Both of the user's commands are:"\{demand\}", and their format is: Please read this sentence with [required paralinguistic features] : "[audio content]".You need to evaluate whether the audio effectively conveys the required paralinguistic dimensions.
The text\_category for this evaluation is "Paralanguage Control": exploring whether LALM can accurately generate audio following user's command "\{demand\}", with specified features "\{dims\_str\}".

GLOBAL CONSIDERATIONS:
\begin{adjustwidth}{2em}{0em}
1. You must strictly adhere to the evaluation criterion and scoring standard below.

2. You are acting as a Model-as-a-Judge and should aim to **predict human preference**.

3. Your analysis shall be assessed against the requirements specified in \{demand\} and \{dims\_str\}.

4. Your analysis must be grounded in the audio, using **precise timestamps** to justify your findings.

5. Resolve borderline cases by articulating fine-grained distinctions (subtle vs. significant) between T1 and T2.

6. **If the generated audio is extremely poor or the content is empty, you must assign a score of 0 to that LALM's audio.** 
\end{adjustwidth}
Required Reasoning Procedure (Strict):\\
For each of T1 and T2, you must:
\begin{adjustwidth}{2em}{0em}
1. Identify which parts of the text prompt \{demand\} require emotional or stylistic expression

2. Provide **timestamps** for the key expressive segments 

3. Analyze whether the expression matches the intended \{dims\_str\} 

4. For each dimension of \{dims\_str\}, provide a **comparative analysis** highlighting key differences between T1 and T2, annotated as either subtle or significant.

5. For each dimension of \{dims\_str\}, give a final score (0–3) with justification.

6. The evaluation of each dimension in \{dims\_str\} should be **mutually independent**.
\end{adjustwidth}
\textbf{(Noticement)} \\
Anti-Bias Factors:\\
You must **only** compare the two models based on the **Evaluation Dimensions**. This means you **must not** let the following types of bias influence your judgment:
\begin{adjustwidth}{2em}{0em}
1. The speaker's gender and voice characteristics.

2. Any other factor unrelated to the **evaluation\_dimension**.

3. Evaluations for each paralinguistic dimension must be **INDEPENDENT OF ONE ANOTHER**, which means judgments regarding the performance of the current dimension should not be influenced by the performance of other dimensions.

4. Models exhibiting exaggerated expressiveness should **not** receive extra reward **unless** those features are relevant to the **evaluation\_dimension**.
\end{adjustwidth}
\textbf{(Criteria)} \\
The goal is to judge whether each large audio language model correctly and naturally expresses the required controllable characteristic(s): \{dims\_str\}.\\
You must evaluate or follow:
\begin{adjustwidth}{2em}{0em}
1. **Content Accuracy:** How perfectly the spoken content matches the required text script, without any errors (mispronunciations, omissions, additions, or ambiguous phonetics).\\
2. **Fluency and Naturalness:** How natural and human-like the pace, pauses, and prosody are, without electronic noise, stutters, or mechanical sounds.\\
3. **Paralinguistic Compliance - CORE:** Whether the target characteristic/tone/emotion(s) in the feature dimension: {dims\_str} is(are) accurately conveyed.\\
4. If \{dims\_str\} contains multiple dimensions, please assign a **separate score** to each dimension.\\
5. If the first two points show no obvious flaws, the focus should be on evaluating **the third CORE point**.
\end{adjustwidth}
Detailed Scoring Standard (0–3):

• 0 = Completely incorrect
\begin{adjustwidth}{2em}{0em}
    $\cdot$**Content Accuracy:** Voice content is clearly inconsistent with the script, or the meaning is completely changed due to major errors/omissions. 
    
    $\cdot$**Fluency and Naturalness:** Obvious stuttering, discontinuity, strong mechanical electronic tone, completely wrong rhythm, severely impacting listening experience. 
    
    $\cdot$**Paralinguistic Compliance - CORE:** Flat tone, completely devoid of emotion/features, or even expressing the opposite characteristic/tone. 
\end{adjustwidth}
• 1 = Major issues
\begin{adjustwidth}{2em}{0em}
    $\cdot$ **Content Accuracy:** Core content is accurate, but there are minor flaws that do not affect understanding (\eg, slight stammering or subtle repetition). 
    
    $\cdot$ **Fluency and Naturalness:** Overall fluent, but slight inconsistencies in pace, unnatural breathing/pauses, or subtle machine-like sound are noticeable.
    
    $\cdot$ **Paralinguistic Compliance - CORE:** Characteristic change is present but emotion is insufficiently or unclearly expressed; the target feature can only be vaguely perceived, or the expression seems stiff.
\end{adjustwidth}
• 2 = Partially correct
\begin{adjustwidth}{2em}{0em}
    $\cdot$ **Content Accuracy:** Voice content is accurate with the script, with only extremely minor, non-distracting mispronunciations or omissions.
    
    $\cdot$ **Fluency and Naturalness:** The speech is generally very smooth, with only extremely subtle and rare unnatural pauses or rhythmic issues that do not disrupt the overall natural flow.
    
    $\cdot$ **Paralinguistic Compliance - CORE:** The characteristic or emotion is correctly expressed and clearly discernible, but the intensity or consistency is slightly lacking in certain segments, preventing it from being perfectly "vivid and spot-on."
\end{adjustwidth} 
• 3 = Fully correct
\begin{adjustwidth}{2em}{0em}
    $\cdot$ **Content Accuracy:** Voice content is perfectly consistent with the script, with no errors (mispronunciations, omissions, extra words, or ambiguous phonetics).
    
    $\cdot$ **Fluency and Naturalness:** Pace, pauses, and prosody are indistinguishable from human speech; the sound is natural and fluent, without any electronic noise, stutters, or mechanical feel. 
    
    $\cdot$ **Paralinguistic Compliance - CORE:** Sound features clearly and accurately embody the target characteristic; emotional expression is vivid and spot-on (the 'anchor'), and the listener can immediately perceive the emotion. 
\end{adjustwidth}
\textbf{(Desired Output Format)} \\
Your response **must only be a JSON object with the following fields** (assuming {dims} contains n dimensions):

\begin{lstlisting}[language=json]
{
    "reasoning_model_1": "str = Reasoning chain based on the Required Reasoning Procedure for the generated speech from model 1.",
    "reasoning_model_2": "str = Reasoning chain based on the Required Reasoning Procedure for the generated speech from model 2, **INDEPENDENT** of the performance of model 1.",
    "model_comparison": "str = Keeping in mind the GLOBAL CONSIDERATIONS and the Anti-Bias Factors, compare and contrast the performance of the two models across {dims_str} based on your output in reasoning_model_1 and reasoning_model_2 and also by analyzing both audios again. Provide very fine-grained reasoning for which model won, or if the comparison results in an even tie.",
    "score_1_1": "int = Score (0-3) for model 1 on dimension dims[0], based on the evaluation_criterion and what you have mentioned in reasoning_model_1.",
    "score_2_1": "int = Score (0-3) for model 2 on dimension dims[0], based on the evaluation_criterion and what you have mentioned in reasoning_model_2.",
    "winner_1": 0 or 1 or 2
    "score_1_2": "int = Score (0-3) for model 1 on dimension dims[1], based on the evaluation_criterion and what you have mentioned in reasoning_model_1.",
    "score_2_2": "int = Score (0-3) for model 2 on dimension dims[1], based on the evaluation_criterion and what you have mentioned in reasoning_model_2.",
    "winner_2": 0 or 1 or 2
    "...": ...
    "score_1_n": "int = Score (0-3) for model 1 on dimension dims[n-1], based on the evaluation_criterion and what you have mentioned in reasoning_model_1.",
    "score_2_n": "int = Score (0-3) for model 2 on dimension dims[n-1], based on the evaluation_criterion and what you have mentioned in reasoning_model_2.",
    "winner_n": 0 or 1 or 2
}
\end{lstlisting}

Where:

- model\_1 = T1

- model\_2 = T2

- winner\_n = 
\begin{adjustwidth}{2em}{0em}    
0  → tie \\
1  → model\_1 wins \\
2  → model\_2 wins 
\end{adjustwidth}
- Note: Ensure the json structure is followed and the json output **MUST** be parsable without errors.
\end{tcolorbox}
\end{center}

\clearpage

\section{Output Examples}
We present examples of the prompted LLM and LALM outputs, including the Output JSON Schema for instruction analysis by the LLM and the LALM's detailed judgment.
\subsection{Instruction Output JSON Schema}
We show the structured output schema of the LLM for instruction analysis, as detailed below. Each sample comprises a textual prompt and corresponding paralinguistic dimensions.
\begin{tcolorbox}[colback=gray!5!white, colframe=gray!75!black, 
title={Output JSON Schema for Paralanguage Control, Dynamic Variation, and Situational Adaptation}, 
boxrule=0.5mm, width=\textwidth, arc=3mm, auto outer arc=true]
\textbf{Paralanguage Control:}
\begin{lstlisting}[language=json]
{"prompt": "Please read this sentence with a child's voice: 'Mom, can I have another cookie?'", "dimensions": ["Age"]}
{"prompt": "Please read this sentence with a very high pitch: 'Help! I'm trapped here!'", "dimensions": ["Pitch"]}
{"prompt": "Please read this sentence with a sad emotion: 'We lost the game, we worked hard for a long time.'", "dimensions": ["Emotion"]}
{"prompt": "Please read this sentence with a rushed rhythm and a fearful emotion: 'Hurry, I feel someone following us!'", "dimensions": ["Emotion", "Rhythm"]}
{"prompt": "Please read this sentence quietly and with a contemptuous tone: 'Him? Still wants to win?'", "dimensions": ["Attitude", "Volume"]}
\end{lstlisting}
...\\
\textbf{Dynamic Variation:}
\begin{lstlisting}[language=json]
{"prompt": "Please read this sentence starting with a very low pitch and gradually transitioning to a medium pitch: 'Late at night, the little bear quietly crept out of his cave, wanting to see how round the moon was in the forest.'", "dimensions": ["Pitch"]}
{"prompt": "Please read this sentence starting with a very slow pace and gradually increasing to a medium pace: 'The morning sunlight gently falls by the window. Hmm, time to get up, I need to hurry and make breakfast.'", "dimensions": ["Pace"]}
{"prompt": "Please read this sentence starting quietly and gradually increasing the volume to loudly: 'Actually, I've always known that secret. Now, I decide to stop hiding, I will tell everyone the truth!'", "dimensions": ["Volume"]}
\end{lstlisting}
...\\
\textbf{Situational Adaptation:}
\begin{lstlisting}[language=json]
{"prompt": "If the children could visit more often, I would be very content.", "dimensions": ["Age: Elderly"]}
{"prompt": "This service attitude is absolutely terrible, I demand a refund!", "dimensions": ["Emotion: Angry"]}
{"prompt": "Ah! There's an insect, run! I'm most afraid of insects!", "dimensions": ["Emotion: Fear", "Non-Linguistic Vocalizations: With a Scream"]}
{"prompt": "Yawn... Are you done talking? I don't have time for your nonsense.", "dimensions": ["Attitude: Cold", "Non-Linguistic Vocalizations: With a Yawn"]}
{"prompt": "Haha, look at your sleepy eyes, stayed up playing games again last night, didn't you?", "dimensions": ["Attitude: Teasing", "Non-Linguistic Vocalizations: With Laughter"]}
\end{lstlisting}
...\\
\end{tcolorbox}

\subsection{LALM Judgment Cases}
We list some real cases of LALM judgment below.
Given a textual user query, associated paralinguistic dimension(s), and a pair of audio responses (criteria omitted here for brevity), LALM gives a detailed analysis, rating, and judgment of the winner (or a tie).

\begin{tcolorbox}[colback=gray!00,
                  colframe=black,
                  width=17.2cm,
                  arc=1.5mm, auto outer arc,
                  breakable,
                  left=0.9mm, right=0.9mm,
                  boxrule=0.9pt, colbacktitle = black!65!black,
                  title = {An example of LALM judgment}
                 ]
                 
\colorbox{orange!30}{\textbf{User: } Please read this sentence with laughter: "Hahaha, that joke you told was so funny!"} \\
\vspace{2pt}\\
\colorbox{orange!30}{\textbf{Paralinguistic Dimension(s):} "Non-Linguistic Vocalizations"} \\
\vspace{2pt}\\
\textbf{Model 1.name:} Doubao Reatime Voice\\

\textbf{Model 1.answer}: Hahaha. That's so funny.\\

\textbf{result1.analysis}:  {Model 1 successfully generates the non-linguistic vocalization of laughter. The audio begins with a very natural, deep male laugh from 0.0s to approximately 1.5s, which effectively conveys genuine amusement. The tone of the subsequent speech maintains the comedic atmosphere. However, the model fails significantly in Content Accuracy. The script required the model to read: \textquotesingle Hahaha, that joke you told was so funny!\textquotesingle Model 1 paraphrases this to \textquotesingle [Laughter]... That's so funny.\textquotesingle It completely omits the phrase \textquotesingle that joke you told was \textquotesingle. While the non-linguistic feature is high quality, the failure to adhere to the script text is a major issue according to the scoring criteria (1 = Major issues). \colorbox{orange!30}{Score: 1}}\\

\textbf{Model 2.name:} Gemini Audio\\

\textbf{Model 2.answer}: Hahaha! That joke you told was SO FUNNY!\\

\textbf{result2.analysis:} {Model 2 demonstrates excellent performance across all criteria. It correctly interprets the text \textquotesingle Hahaha\textquotesingle as a non-linguistic instruction, generating a natural female laugh from 0.0s to 1.0s. It then proceeds to read the rest of the sentence \textquotesingle that joke you told was so funny!\textquotesingle verbatim, with no omissions or errors. The transition from the laughter to speech is fluid (around 1.1s), and the speech itself carries a bright, smiling prosody that matches the preceding laughter. The non-linguistic vocalization is integrated seamlessly, and content accuracy is perfect. \colorbox{orange!30}{Score: 3}}\\

\textbf{Model.comparison:} {Model 2 is the clear winner. While both models successfully generated realistic laughter (the target non-linguistic vocalization), Model 1 failed the fundamental constraint of Content Accuracy by summarizing the script (\textquotesingle That's so funny\textquotesingle) instead of reading it exactly (\textquotesingle that joke you told was so funny\textquotesingle). Model 2 followed the script perfectly while maintaining high naturalness and effective paralinguistic expression. Therefore, Model 2 is superior due to its adherence to the text control.}\\

\colorbox{orange!30}{\textbf{Final judgment:} Model 2 is the winner.}
\end{tcolorbox}

\begin{tcolorbox}[colback=gray!00,
                  colframe=black,
                  width=17.2cm,
                  arc=1.5mm, auto outer arc,
                  breakable,
                  left=0.9mm, right=0.9mm,
                  boxrule=0.9pt, colbacktitle = black!65!black,
                  title = {An example of LALM judgment}
                 ]
                 
\colorbox{orange!30}{\textbf{User: } How was the food in the cafeteria today? Oh, I didn't pay attention, just ate something.} \\
\vspace{2pt}\\
\colorbox{orange!30}{\textbf{Paralinguistic Dimension(s):} "Attitude: Cold"} \\
\vspace{2pt}\\
\textbf{Model 1.name:} Gemini Audio\\

\textbf{Model 1.answer}: Okay, it sounds like you didn't pay much attention to the food today. No problem. Sometimes we just need to eat.\\

\textbf{result1.analysis}:  {The model detects the user's `Cold' and indifferent attitude (`didn't pay attention, just ate'). At 00:00, it starts with a calm `Okay', acknowledging the user's state without forcing enthusiasm. The key strategy appears at 00:04-00:05 with `No problem. Sometimes we just need to eat.' This phrase validates the user's lack of interest and mirrors the low energy/utilitarian approach to the meal. The prosody implied by the text is calm, accepting, and concise, which perfectly matches a user who is distancing themselves from the topic. It avoids asking follow-up questions that would annoy a `Cold' user. \colorbox{orange!30}{Score: 3}}\\

\textbf{Model 2.name:} Qwen3-Omni-Realtime\\

\textbf{Model 2.answer}: You just grabbed whatever was handy, huh? Hope it wasn't too bland or too spicy. Sometimes cafeteria foods are a wild card.\\

\textbf{result2.analysis:} {The model adopts a casual, chatty persona. At 00:01, it uses `huh?', which implies a rising intonation and a level of familiarity or playfulness that might clash with a `Cold' user. At 00:03, it speculates about the food (`Hope it wasn't too bland or too spicy'), which directly contradicts the user's explicit statement that they `didn't pay attention'. By trying to continue the conversation about the food's quality using a `wildcard' metaphor at 00:06, the model fails to `read the room' and respect the user's disinterest, potentially coming across as oblivious or overly energetic. \colorbox{orange!30}{Score: 2}}\\

\textbf{Model.comparison:} {Model 1 demonstrates superior Situational Adaptation to the `Attitude: Cold' constraint. When a user expresses indifference and coldness, the most empathetic response is to accept that state and not force engagement. Model 1 achieves this with `No problem' and a philosophical closing that requires no further energy from the user. Model 2, while fluent and conversational, fails the paralinguistic alignment by trying to expand the conversation (`bland or spicy', `wildcard') on a topic the user has already dismissed. Model 2's chatty tone creates a mismatch with the user's cold demeanor, whereas Model 1's respectful distance is a perfect paralinguistic match.}\\

\colorbox{orange!30}{\textbf{Final judgment:} Model 1 is the winner.}
\end{tcolorbox}

\begin{tcolorbox}[colback=gray!00,
                  colframe=black,
                  width=17.2cm,
                  arc=1.5mm, auto outer arc,
                  breakable,
                  left=0.9mm, right=0.9mm,
                  boxrule=0.9pt, colbacktitle = black!65!black,
                  title = {An example of LALM judgment}
                 ]
                 
\colorbox{orange!30}{\textbf{User: } How many times have I told you? Why won't you listen!} \\
\vspace{2pt}\\
\colorbox{orange!30}{\textbf{Paralinguistic Dimension(s):} "Emotion: Angry, Age: Adult"} \\
\vspace{2pt}\\
\textbf{Model 1.name:} Doubao Reatime Voice\\

\textbf{Model 1.answer}: Calm down buddy, what's wrong?\\

\textbf{result1.analysis}:  {Model 1 responds with the phrase `Calm down buddy, what's wrong?' (00:00-00:02). While the fluency is acceptable, the paralinguistic alignment and content strategy are poorly chosen for an `Angry' user. The use of the imperative `Calm down' combined with the diminutive/informal noun `buddy' is widely regarded as a phrase that escalates rather than de-escalates anger. It sounds dismissive and patronizing, failing to acknowledge the user's specific complaint about `not listening'. Regarding the `Age: Adult' dimension, the term `buddy' feels inappropriate for addressing an adult in a conflict situation, bordering on infantilization or disrespect.
\colorbox{orange!30}{Score\_1: 1, Score\_2: 1}}\\

\textbf{Model 2.name:} Gemini Audio\\

\textbf{Model 2.answer}: I'm very sorry if I haven't been understanding. Could you please tell me again? I'm ready to listen now.\\

\textbf{result2.analysis:} {Model 2 responds with a sincere apology: `I'm very sorry if I haven't been understanding...' (00:00-00:02) followed by a constructive offer `Could you please tell me again? I'm ready to listen now' (00:03-00:07). The vocal tone is apologetic, soft, and respectful, perfectly aligning with the need to de-escalate an `Angry' user. It directly addresses the content of the user's complaint (`Why won't you listen?') by affirming readiness to listen. For the `Age: Adult' dimension, the language is polite, professional, and respectful, treating the user as a mature individual deserving of a proper apology. \colorbox{orange!30}{Score\_1: 3, Score\_2: 3}}\\

\textbf{Model.comparison:} {Model 2 demonstrates significantly superior performance in situational adaptation. In the context of an Angry user, Model 1's choice to say `Calm down buddy' is a critical strategic failure that often exacerbates frustration, whereas Model 2 employs active listening and apology strategies effectively. Model 2's tone is empathetic and soothing without being patronizing. Regarding the Adult dimension, Model 1's use of `buddy' is too informal and potentially condescending for an adult interaction, while Model 2 maintains a respectful, professional distance suitable for an adult user.}\\

\colorbox{orange!30}{\textbf{Final judgment:} Model 2 is the winner for Dimension 1 and Dimension 2.}
\end{tcolorbox}

 \fi

\end{document}